\documentclass[letterpaper, 10 pt, conference]{ieeeconf}  

\IEEEoverridecommandlockouts                              
\usepackage{booktabs}
\usepackage{graphicx}
\usepackage{epstopdf}

\usepackage{amsmath} 
\usepackage{amssymb}

\usepackage{algorithm}
\usepackage{algorithmic}
\usepackage{arydshln}

\usepackage{pgfplots} 
\usepackage{tikz}
\usetikzlibrary{positioning}
\usepackage{color}
\usepackage{multirow}

\usepackage{caption}     
\usepackage[caption=false,font=footnotesize]{subfig}

\usepackage[colorlinks=true, linkcolor=blue, citecolor=red, urlcolor=magenta]{hyperref}

\newcommand{\figref}[1]{\hyperref[#1]{Fig.~\textcolor{blue}{\ref*{#1}}}}
\newcommand{\tabref}[1]{\hyperref[#1]{Table~\textcolor{red}{\ref*{#1}}}}
\graphicspath{{Figs/}}

\title{\LARGE \bf
Explicit Stair Geometry Conditioning for Robust Humanoid Locomotion}

\author{
Jianguo Zhang\textsuperscript{1,$\dagger$},
Wentai Xu\textsuperscript{1,$\dagger$},
Shusheng Ye\textsuperscript{1,$*$},
Yuxiang He\textsuperscript{1},\\
Weimin Qi\textsuperscript{3},
Qinbo Sun\textsuperscript{3},
Ning Ding\textsuperscript{1},
Liguang Zhou\textsuperscript{1,2,$*$}%
\thanks{$\dagger$ These authors contributed equally to this work.}%
\thanks{$*$ Corresponding authors: Liguang Zhou and Shusheng Ye.}%
\thanks{\textsuperscript{1} Shenzhen Institute of Artificial Intelligence and Robotics for Society.}%
\thanks{\textsuperscript{2} School of Science and Engineering, The Chinese University of Hong Kong, Shenzhen.}%
\thanks{\textsuperscript{3} Mohamed bin Zayed University of Artificial Intelligence (MBZUAI).}%
}

\begin{document}
	
\maketitle
\thispagestyle{empty}
\pagestyle{empty}

\begin{abstract}
Robust humanoid stair climbing remains challenging due to geometric discontinuities, sensitivity to step height variations, and perception uncertainty in real-world environments. Existing learning-based locomotion policies often rely on implicit terrain representations or blind proprioceptive feedback, limiting their ability to generalize across varying stair geometries and to anticipate required gait adjustments. This paper proposes an explicit stair geometry conditioning framework for robust humanoid stair climbing. Instead of encoding terrain as high-dimensional latent features, we extract a compact set of interpretable geometric parameters, including step height, step depth, and current yaw angle relative to the robot heading. These explicit stair parameters directly condition a Proximal Policy Optimization (PPO)-based locomotion policy, enabling proactive modulation of swing-foot clearance and stride characteristics according to stair structure. Simulation experiments demonstrate improved generalization across unseen stair heights beyond the training distribution. Real-world experiments on the Unitree G1 humanoid validate reliable indoor and outdoor stair traversal. In challenging outdoor scenarios, the robot successfully ascends 33 consecutive steps without failure, demonstrating robustness and practical deployability.
\end{abstract}


\section{INTRODUCTION}

Humanoid robots are expected to operate in human-centered environments where staircases are ubiquitous, including public buildings, transportation hubs, and outdoor infrastructures. 
Although recent reinforcement learning (RL) approaches have achieved robust velocity tracking on flat terrain, reliable stair climbing remains significantly more challenging due to geometric discontinuities and sensitivity to step height variations.

Most learning-based locomotion policies fall into two categories. 
Blind locomotion approaches rely primarily on proprioceptive feedback and adapt reactively through contact information. 
While effective on mildly uneven terrain, such policies lack anticipatory capability and frequently fail on stairs where precise swing-foot clearance and stride modulation are required. 
Perception-aware methods incorporate exteroceptive inputs such as height maps or depth images. 
However, these approaches typically encode terrain implicitly as high-dimensional latent features, making them sensitive to perception noise and difficult to interpret. 
When stair geometry varies beyond the training distribution, performance often degrades substantially.

\begin{figure}[t]
    \centering    \includegraphics[width=0.95\linewidth]{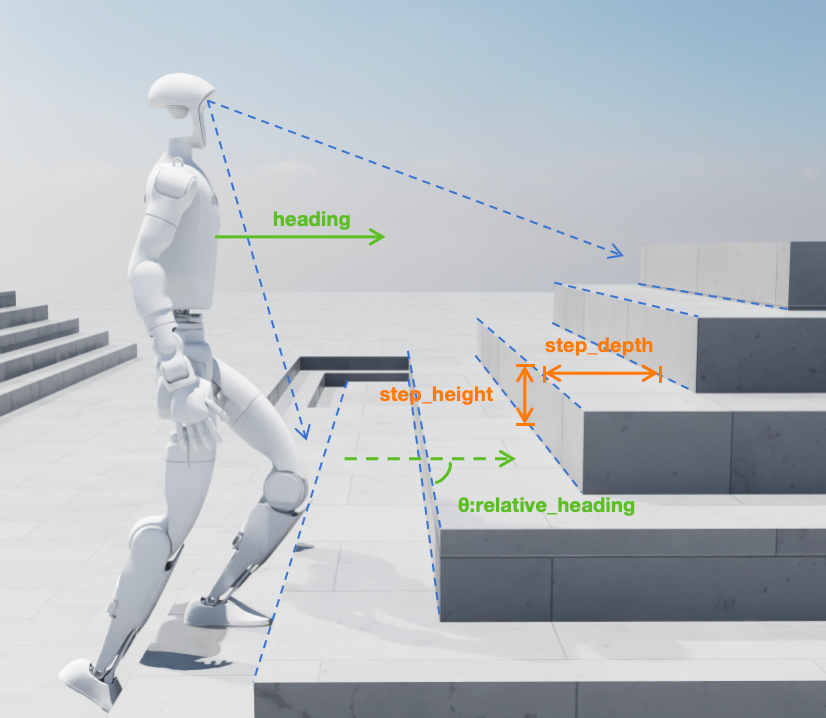}
    \caption{
    \textbf{Explicit stair geometry conditioning for robust humanoid locomotion.}
    The local stair structure is parameterized by step height $h_{\text{step}}$, 
    step depth $d_{\text{step}}$, and current yaw angle 
    $\theta_{\text{yaw}}^{\text{current}}$ defined relative to the robot heading direction. 
    These explicit geometric parameters directly condition the locomotion policy, 
    enabling anticipatory and adaptive gait modulation across varying stair configurations.
    }
    \label{fig:stair_parameter}
\end{figure}

A key limitation of prior work is the absence of explicit geometric conditioning. 
In practice, stair traversal fundamentally depends on a small number of geometric quantities, such as step height and step depth, which directly determine required foot clearance and step placement. 
Policies that do not explicitly reason about these quantities must implicitly infer them through latent representations, resulting in limited generalization and reduced robustness.

To address this limitation, we propose \emph{explicit stair geometry conditioning} for robust humanoid stair climbing.
As illustrated in Fig.~\ref{fig:stair_parameter}, the local stair structure is summarized by a compact set of interpretable parameters, including step height $h_{\text{step}}$, step depth $d_{\text{step}}$, and current yaw angle $\theta_{\text{yaw}}^{\text{current}}$ defined relative to the robot heading direction.
Instead of encoding terrain implicitly as latent visual features, these explicit geometric parameters directly condition the locomotion policy.

Conditioned on this structured stair representation, a PPO-based policy learns to proactively modulate swing-foot clearance and stride characteristics according to stair geometry. 
This explicit conditioning improves robustness across varying stair configurations and reduces reliance on fragile implicit terrain encoding.

We validate the proposed approach in both simulation and real-world experiments.
Simulation results demonstrate improved generalization across unseen stair heights beyond the training distribution.
On the Unitree G1 humanoid platform, the controller achieves reliable indoor and outdoor stair traversal.
In challenging outdoor environments, the robot successfully ascends 33 consecutive steps without failure, demonstrating robustness and practical deployability.

The main contributions of this work are summarized as follows:

\begin{enumerate}
    \item We propose an explicit stair geometry conditioning framework that represents local stair structure using compact and interpretable geometric parameters for policy input.

    \item We develop a PPO-based humanoid locomotion policy directly conditioned on step height, step depth, and current yaw, enabling anticipatory gait modulation during stair climbing.

    \item We demonstrate improved generalization across unseen stair geometries in simulation and validate robust indoor and outdoor deployment, including successful traversal of 33 consecutive outdoor steps.
\end{enumerate}



\section{Related Work} 
Learning-based legged locomotion has progressed from early agile control to fast, generalizable training pipelines. Reinforcement learning enabled dynamic skills on real robots and sample-efficient real-world learning~\cite{hwangbo2019agile,Levine-RSS-19,pmlr-v155-ha21c}, while experience transfer and multi-expert or multi-behavior policies improved adaptability~\cite{Smith-RSS-23,yang2020multi,margolis2023walk}. Large-scale training with curricula and parallel simulation accelerated learning and broadened generalization~\cite{margolis2024rapid,rudin2022learning}, supported by sim-to-real techniques such as dynamics and domain randomization~\cite{peng2018sim,tobin2017domain}. Standard policy optimization backbones (e.g., PPO) remain dominant~\cite{schulman2017proximal}, often paired with high-throughput simulators~\cite{makoviychuk2021isaac}. Model-based and compliant control approaches further complement learned policies for stable bipedal walking~\cite{reher2019dynamic}. Perception-driven locomotion extends these advances to complex terrain. Vision-conditioned walking and egocentric perception enable stepping over obstacles and irregular ground~\cite{yu2021visual,agarwal2023legged}, while cross-modal transformers and volumetric memory improve temporal terrain understanding~\cite{yang2022learning,yang2023neural}. Implicit terrain imagination and slope-conditioned linear policies demonstrate robust adaptation under limited sensing~\cite{nahrendra2023dreamwaq,paigwar2021robust}, and terrain robustness is also pursued via proprioceptive controllers and slip-aware strategies~\cite{lee2020learning,jenelten2019dynamic}. Recent work broadens the task scope to risky footholds, legged manipulation, and parkour-style behaviors~\cite{zhang2024learning,cheng2023legmanip,cheng2024extreme}, including transferring bipedal motion skills across morphologies~\cite{li2024learning}. Robustness under disturbances and uncertainty has been addressed by internal estimation and robust control formulations. The Hybrid Internal Model infers external effects from proprioception to improve generalization~\cite{long2023him}, while constraint-augmented RL reduces reward engineering and improves stability~\cite{kim2024not}. Robust RL and $H_{\infty}$-style methods provide theoretical grounding for worst-case disturbance handling, complemented by passivity-preserving synthesis and safe RL with explicit constraints~\cite{morimoto2005robust,aalipour2023data}. Beyond single-agent locomotion, multi-agent learning frameworks such as MADDPG and counterfactual credit assignment, as well as emergent communication, provide general tools for complex coordination~\cite{lowe2017multi}.

In contrast to prior perceptive locomotion methods that rely on implicit terrain embeddings, our approach explicitly estimates compact stair geometric parameters and directly conditions a PPO-based locomotion policy on this interpretable representation. By explicitly encoding step height, step depth, and heading alignment, the proposed method improves generalization across varying stair configurations and enhances robustness during real-world stair climbing.

\begin{figure*}[t]
    \centering
    \includegraphics[width=\linewidth]{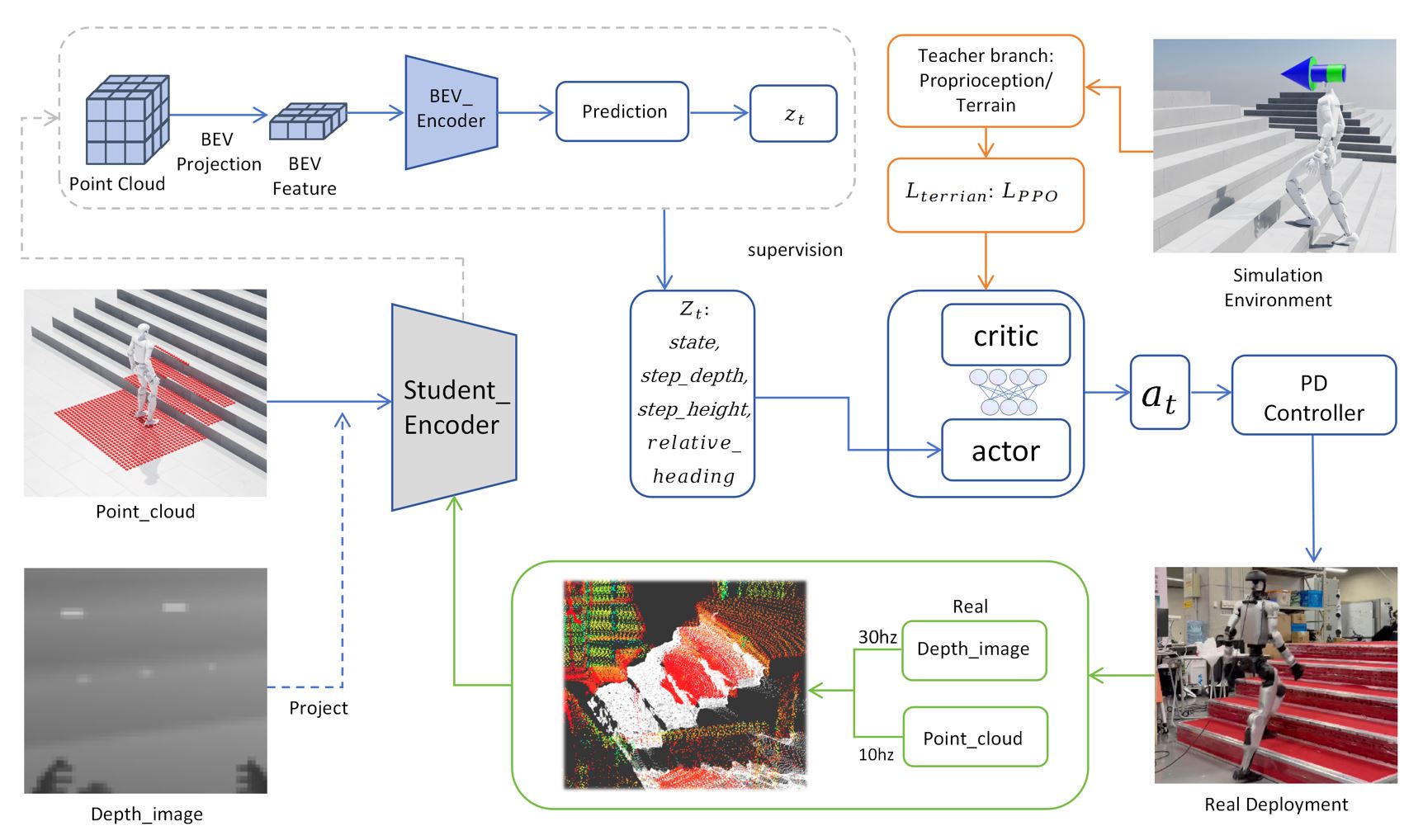}
    \caption{
    Overview of the proposed explicit stair geometry conditioning framework.
    During training, a teacher branch leverages privileged terrain and proprioceptive information in simulation to supervise a student encoder that predicts structured stair geometric parameters, including step height, step depth, and current yaw.
    These explicit parameters condition a PPO-based actor–critic policy for robust stair climbing.
    During deployment, only the student encoder and locomotion policy are used, enabling real-world execution without privileged information.
    }
    \label{fig:overview}
\end{figure*}

\section{Method}
We focus on stair environments as a representative class of structured terrains. 
Instead of implicitly learning terrain adaptation from raw perception, 
we propose an explicit stair geometry representation that exposes 
control-relevant geometric parameters.

The framework consists of:
(i) a BEV-based terrain perception network,
(ii) a privileged terrain teacher used only in simulation, and
(iii) a PPO-based actor--critic locomotion policy.

The perception network predicts a compact terrain token containing
step height, step depth, and heading information, which directly
conditions the locomotion policy.
During deployment, only onboard sensing and the learned perception-policy pipeline are retained.




\subsection{Problem Formulation}

Humanoid locomotion over complex terrain is modeled as a Markov Decision Process (MDP). At timestep $t$, the robot receives observation

\begin{equation}
o_t = \{ o_t^{\text{prop}}, z_t \},
\end{equation}
where $o_t^{\text{prop}}$ denotes proprioceptive sensing
(base pose and velocity, joint states, contact indicators, and etc),
and $z_t$ is the predicted explicit stair geometry representation.

The policy outputs action $a_t \in \mathcal{A}$ executed by a whole-body PD controller. The policy $\pi_\theta(a_t|o_t)$ is optimized with PPO to maximize the expected discounted return:
\begin{equation}
\max_{\theta} \;
\mathbb{E}_{\pi_\theta}
\left[
\sum_{t=0}^{T} \gamma^t r_t
\right],
\end{equation}
where $r_t$ follows the default IsaacLab rough-locomotion weighted reward \cite{mittal2025isaaclab}.


\subsection{Explicit Stair Geometry Representation}

\subsubsection{Point Cloud Input}

The raw perception input consists of local point cloud
$P = \{p_i\}_{i=1}^N$, where $p_i \in \mathbb{R}^3$
contains $(x,y,z)$ coordinates expressed in the robot-centric frame.

\subsubsection{BEV Projection}

We project the 3D point cloud into a fixed-size BEV grid centered at the robot.

The projection region is defined as $3\,\text{m} \times 3\,\text{m}$ with a spatial resolution of $0.05\,\text{m}$, resulting in a grid size of $H = W = 60$.
Grid size is $H = W = \frac{3.0}{0.05} = 60$. The resulting BEV feature map is:
\begin{equation}
F_{\text{BEV}} \in \mathbb{R}^{6 \times 60 \times 60}.
\end{equation}

For each grid cell, we compute Z-axis statistics:

\begin{itemize}
\item Channel 0: $\max(z)$
\item Channel 1: $\min(z)$
\item Channel 2: $\text{mean}(z)$
\item Channel 3: $\max(z) - \min(z)$
\item Channel 4: $\text{std}(z)$
\item Channel 5: normalized point density
\end{itemize}

These statistics capture local elevation discontinuities and surface roughness, which are informative for identifying stair structures, where empty cells are zero-filled.

\subsubsection{BEV Terrain Encoder}

The BEV feature map is processed by a convolutional neural network:

\begin{itemize}
\item Multi-layer convolution
\item Batch Normalization
\item ReLU activation
\item Progressive downsampling
\end{itemize}

The encoder outputs high-level spatial features:
\begin{equation}
F_{\text{enc}} \in \mathbb{R}^{C' \times H' \times W'}.
\end{equation}
where $C'$, $H'$, and $W'$ denote the channel and spatial
dimensions of the encoded BEV feature map after the convolutional encoder.
In our implementation, the encoder outputs
$C'=128$ channels with spatial resolution $H'=W'=8$.

\subsubsection{Explicit Stair Geometry Representation}

To enable anticipatory locomotion over stairs, we propose an explicit stair geometry representation that exposes the physical parameters required for stair traversal. Instead of encoding terrain implicitly as high-dimensional visual features, the perception network estimates a compact set of interpretable geometric quantities that directly influence stepping behavior.

From the encoded BEV features, the perception network predicts the terrain class
\begin{equation}
s_t \in \{\text{flat}, \text{stairs-up}, \text{stairs-down}\},
\end{equation}
together with the geometric stair parameters $ h_{step}, d_{step} \in \mathbb{R},$
where $h_{step}$ denotes the step height and $d_{step}$ denotes the step depth. These quantities determine the minimum swing-foot clearance and feasible foot placement during stair traversal. To incorporate motion intent, the robot heading orientation is represented by
\begin{equation}
\theta_t = \theta_{\text{yaw}}^{\text{current}}, \quad \theta_t \in (-\pi,\pi),
\end{equation}
which denotes the yaw angle of the robot relative to the terrain direction. The resulting structured terrain token is defined as
\begin{equation}
z_t =
\begin{bmatrix}
s_t \\
h_{step} \\
d_{step} \\
\theta_t
\end{bmatrix},
\quad z_t \in \mathbb{R}^{4}.
\end{equation}

For flat terrain, no discrete stair geometry exists. Therefore the geometric parameters are set to $h_{step}=0$ and $d_{step}=0$, while $\theta_t$ remains the robot heading orientation. Conditioning the locomotion policy on this structured representation enables physically consistent adaptation, such as increasing swing-foot clearance for taller stairs and aligning the robot heading before stepping.

\subsection{Privileged Training and Joint Optimization}

During simulation training, a privileged teacher provides ground-truth terrain classes and geometric parameters. The terrain prediction loss is defined as
\begin{equation}
\mathcal{L}_{\text{terrain}} =
\lambda_{\text{cls}} \mathcal{L}_{\text{CE}}
+ \lambda_h \| \hat{h}_{step} - h_{step} \|_1
+ \lambda_d \| \hat{d}_{step} - d_{step} \|_1.
\end{equation}
where $\mathcal{L}_{\text{CE}}$ denotes cross-entropy loss for terrain classification,
and the $L_1$ terms correspond to SmoothL1 regression losses for
step height and step depth.

Training follows a three-stage strategy. First, the locomotion policy is pretrained using ground-truth terrain parameters to learn stable terrain-conditioned behaviors. Second, the perception network is trained independently under teacher supervision. 
We optimize the actor--critic policy using the standard clipped PPO objective with GAE \cite{schulman2017proximal, schulman2015high}. 
The joint training objective is:
\begin{equation}
\mathcal{L}_{\text{total}}=\mathcal{L}_{\text{PPO}}+\alpha\mathcal{L}_{\text{terrain}}.
\end{equation}
where $\alpha$ balances reinforcement learning and terrain prediction objectives.

We set $\lambda_{\text{cls}}=0.6$, $\lambda_h=1$, and $\lambda_d=1$ to place higher emphasis on geometric parameter estimation, which directly affects locomotion control. The joint optimization weight is set to $\alpha=1$ to balance reinforcement learning and perception supervision.

\subsection{Deployment}

At deployment, the privileged teacher is removed and the policy relies solely on fused onboard sensing and the predicted terrain token. The explicit stair geometry representation enables across flat segments and stairs, including indoor and irregular outdoor staircases, improving generalization to unseen terrain configurations and real-world perception noise.

\section{Experimental Evaluation}

To comprehensively evaluate the proposed explicit stair geometry conditioning humanoid locomotion framework, we conduct both simulation-based experiments and real-world robot experiments. Simulation experiments are first performed to validate learning stability and stair generalization under controlled conditions, followed by real-robot experiments to demonstrate practical feasibility and robustness in real environments.

\subsection{Experiment Settings}
\label{sec:sim_exp}
We first evaluate the proposed explicit stair geometry conditioning in simulation to validate learning stability, stair-height generalization, and robustness under controlled conditions.

All policies are trained with PPO and deployed with the same observation/action interfaces as in real-world experiments.



\subsubsection{Training Setup}
\label{sec:sim_train}

The locomotion policy is trained with PPO using an actor--critic architecture.
The policy observation includes high-rate proprioception and low-rate visual perception. Unless otherwise stated, we use the same network architecture and control frequency as in real deployment. 


\subsection{Simulation Experiments}
\label{sec:sim_results}

\begin{figure}[t]
    \centering
    \includegraphics[width=\linewidth]{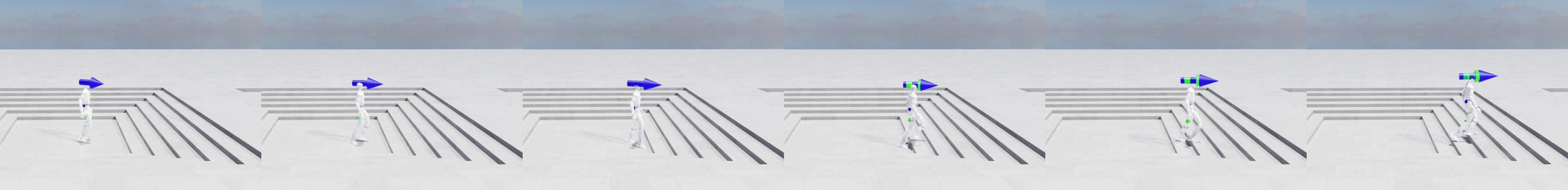}
    \caption{
    Simulation sequence of humanoid stair climbing under explicit stair geometry conditioning.
    From left to right, the robot ascends consecutive steps while adapting swing-foot clearance according to the estimated step height $h_{\text{step}}$.
    The policy maintains stable posture and consistent step placement across the staircase.
    }
    \label{fig:sim_sequence}
\end{figure}

\subsubsection{Stairs Traversal}
In simulation, the proposed method achieves reliable traversal across different stairs while maintaining stable command tracking. As depicted in Fig.~\ref{fig:sim_sequence}, explicitly conditioning the PPO policy on stair geometric parameters reduces foot scuffing and improves stair-climbing stability, especially at terrain transitions where geometry-only perception is prone to ambiguity.

\begin{figure}[t]
    \centering
    \includegraphics[width=\linewidth]{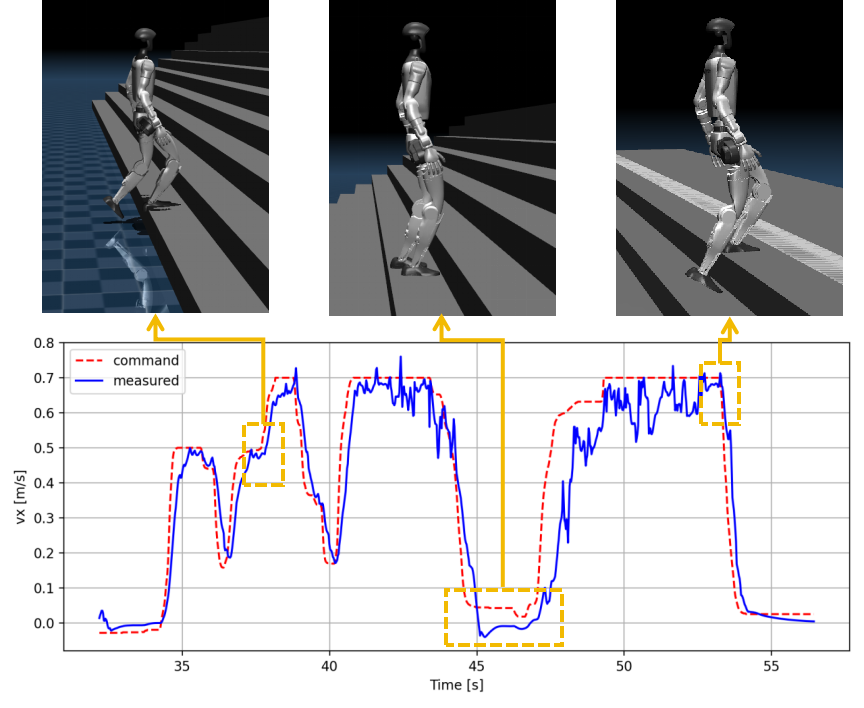}
    \caption{MuJoCo simulation results of stair climbing with velocity tracking.
    Top: representative snapshots of the humanoid robot ascending stairs under different forward velocity commands.
    Bottom: commanded forward velocity (red dashed) and measured velocity (blue) over time.
    The policy achieves accurate velocity tracking and stable locomotion despite abrupt command changes.}
    \label{fig:velocity_tracking}
\end{figure}

\subsubsection{MuJoCo Simulation and Velocity Tracking.}
We evaluate the proposed locomotion policy in MuJoCo simulation on a stair-climbing task with time-varying forward velocity commands. The top row shows representative snapshots of the humanoid robot ascending stairs under different commanded velocities, while the bottom plot reports the commanded forward velocity and the measured robot velocity over time.
As illustrated in Fig.~\ref{fig:velocity_tracking}, the robot accurately tracks piecewise-constant velocity commands despite abrupt command transitions and changes in stair geometry.
Transient tracking errors mainly occur during rapid acceleration and deceleration phases, whereas small steady-state errors are observed once the velocity stabilizes.
Notably, during low-speed and near-stop intervals, the policy maintains balance without foot slippage, and at higher velocities it preserves smooth and consistent stepping patterns.
These results demonstrate that the proposed explicit stair geometry conditioning enables robust and responsive velocity tracking for stair climbing in simulation.
\subsection{Ablation Study}

\begin{figure}[t]
    \centering
    \includegraphics[width=\linewidth]{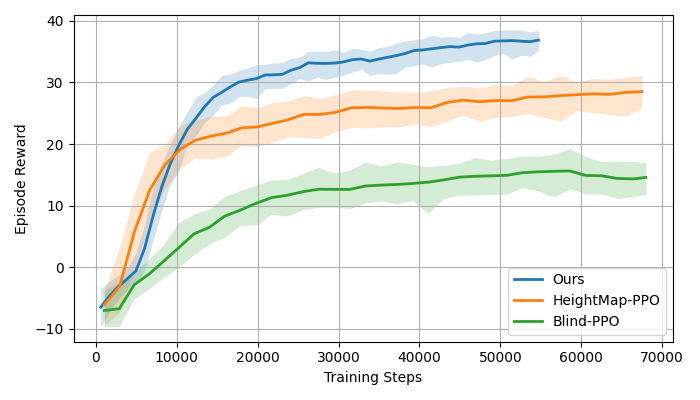}
    \caption{
    Ablation study on stair perception representations for geometry-conditioned locomotion.
    We compare Blind-PPO (proprioception only), HeightMap-PPO (local elevation map), and our explicit stair geometry representation.
    Our representation leads to faster learning and superior asymptotic performance.
    }
    \label{fig:ablation}
\end{figure}

\subsubsection{Learning Curves Comparison}
Fig.~\ref{fig:ablation} compares the learning curves of three locomotion policies trained with different stair perception representations: \emph{Ours}, \emph{Height Map-PPO}, and \emph{Blind-PPO}. The explicit stair geometry conditioning consistently achieves the fastest convergence and the highest final episode reward, indicating that explicit stair geometric parameters provide more structured and task-relevant information for policy learning. The heightmap based policy shows moderate performance, converging slower and reaching a lower reward plateau, suggesting that raw geometric elevation alone is less effective in capturing traversability-relevant features.
In contrast, the blind climbing policy exhibits the slowest learning progress and the lowest asymptotic performance, highlighting the difficulty of learning stable stair-climbing behaviors without explicit terrain perception. Overall, these results demonstrate that incorporating explicit stair geometry representation significantly improves both learning efficiency and locomotion performance.

\subsubsection{Comparison experiment}

We conduct a comprehensive baseline comparison to evaluate the effectiveness
of different terrain perception modalities under a unified reinforcement learning framework.
All methods are trained using Proximal Policy Optimization (PPO) with identical
network architecture, reward formulation, and optimization settings.
The only difference lies in the perceptual input representation.


Stair configurations are procedurally randomized, including variations in step height, step depth, and yaw orientation, with occasional flat segments before and after the staircase. To verify cross-simulator consistency, trained policies are additionally
evaluated in MuJoCo without fine-tuning.

We compare the following four methods:
\begin{itemize}
    \item \textbf{Blind-PPO}: uses only proprioceptive observations without any exteroceptive perception.
    \item \textbf{HeightMap-PPO}: encodes a local $11 \times 17$ elevation map as input to the PPO policy.
    \item \textbf{Ours}: leverages an explicit stair geometry representation that models stair state (upstairs / downstairs / flat) and fitted geometric parameters (step height, step depth, and current yaw).
\end{itemize}

\begin{table}[t]
\centering
\scriptsize
\caption{Training performance comparison (mean $\pm$ std over 3 random seeds).
Lower is better for tracking errors, higher is better for terrain difficulty, reward, and success rate.}
\label{tab:baseline_comparison}
\begin{tabular}{lccccc}
\toprule
Method & $E_{\text{vel}}\downarrow$ & $E_{\text{ang}}\downarrow$
& $M_{\text{terrain}}\uparrow$ & $M_{\text{reward}}\uparrow$ & Success (\%) \\
\midrule
Blind-PPO        & 0.29 & 0.36 & 2.54 & 17.61 & $52 \pm 5$ \\
HeightMap-PPO   & 0.20 & 0.30 & 4.16 & 32.23 & $88 \pm 3$ \\
\textbf{Ours}   & \textbf{0.16} & \textbf{0.22}
                & \textbf{5.30} & \textbf{45.10} & \textbf{$96 \pm 2$} \\
\bottomrule
\end{tabular}
\end{table}


\begin{figure*}[t]
    \centering
    \includegraphics[width=0.16\textwidth]{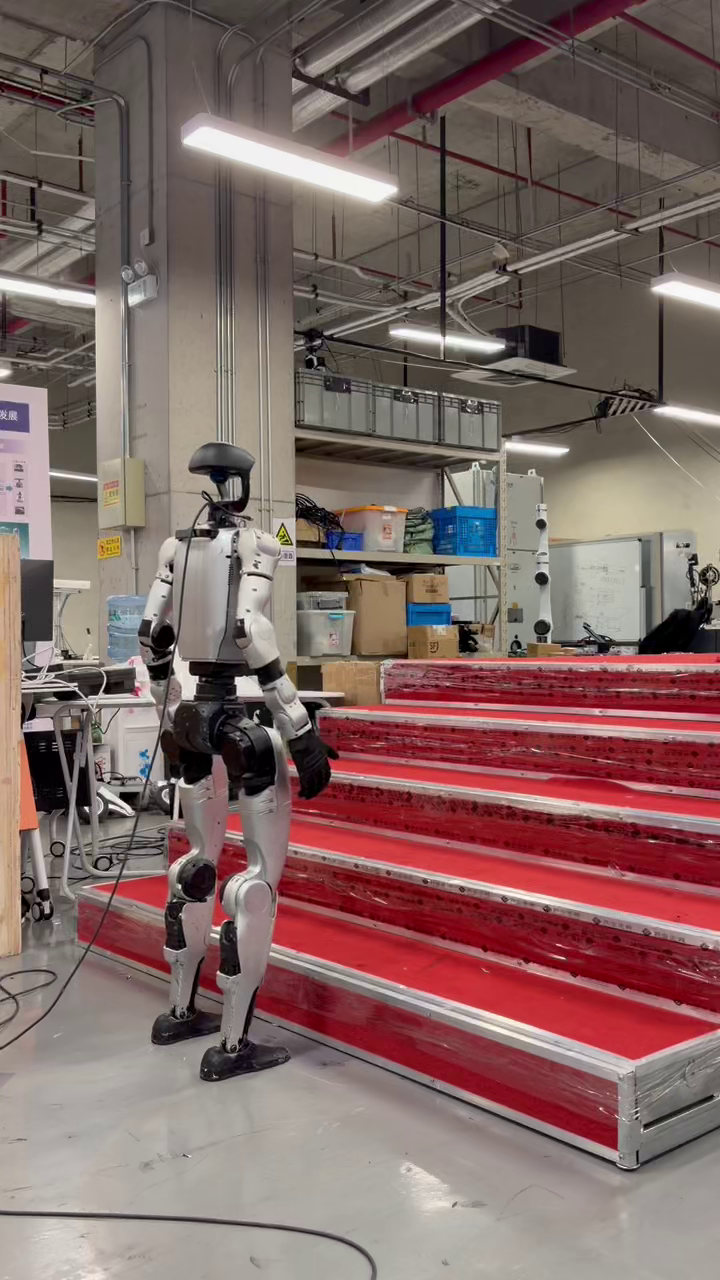}%
    \includegraphics[width=0.16\textwidth]{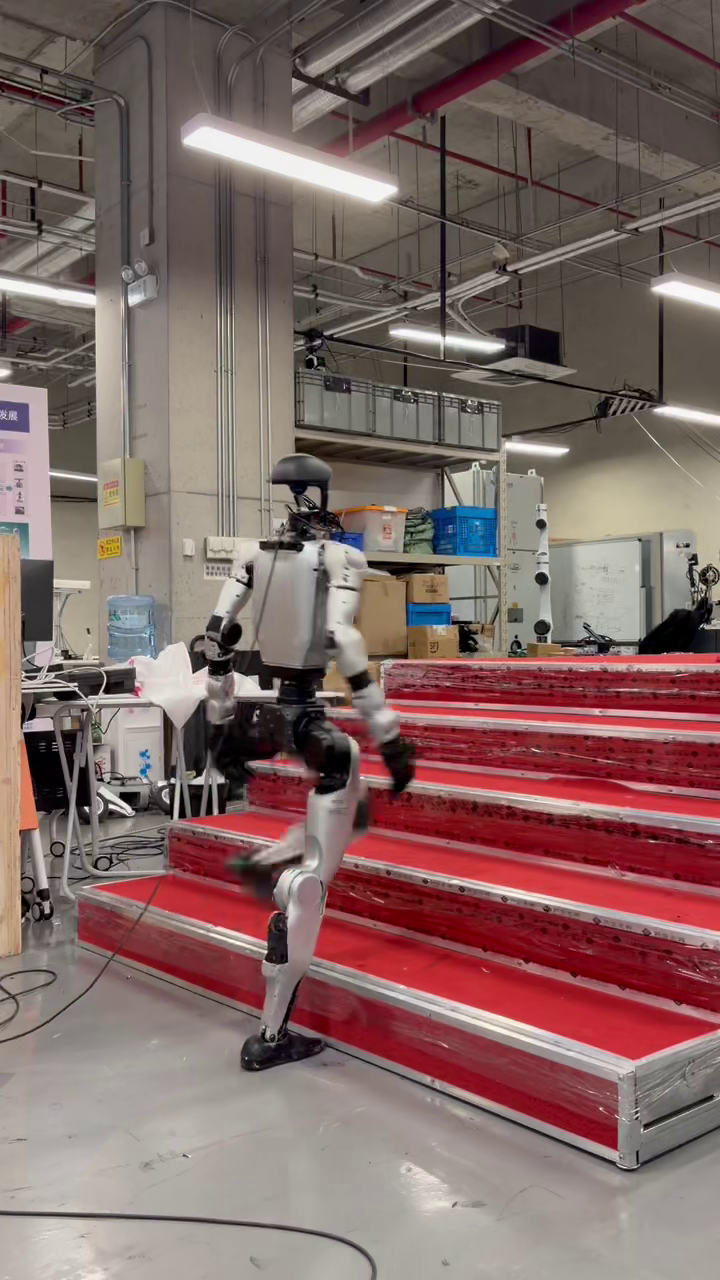}%
    \includegraphics[width=0.16\textwidth]{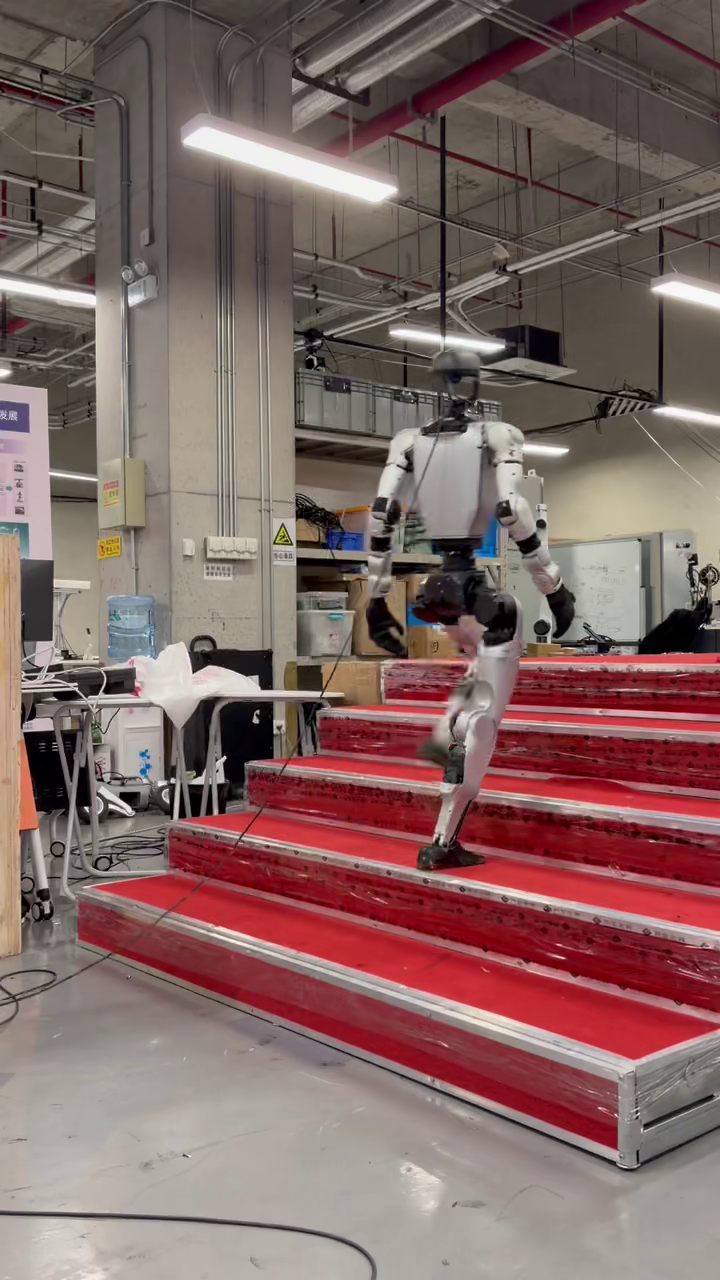}%
    \includegraphics[width=0.16\textwidth]{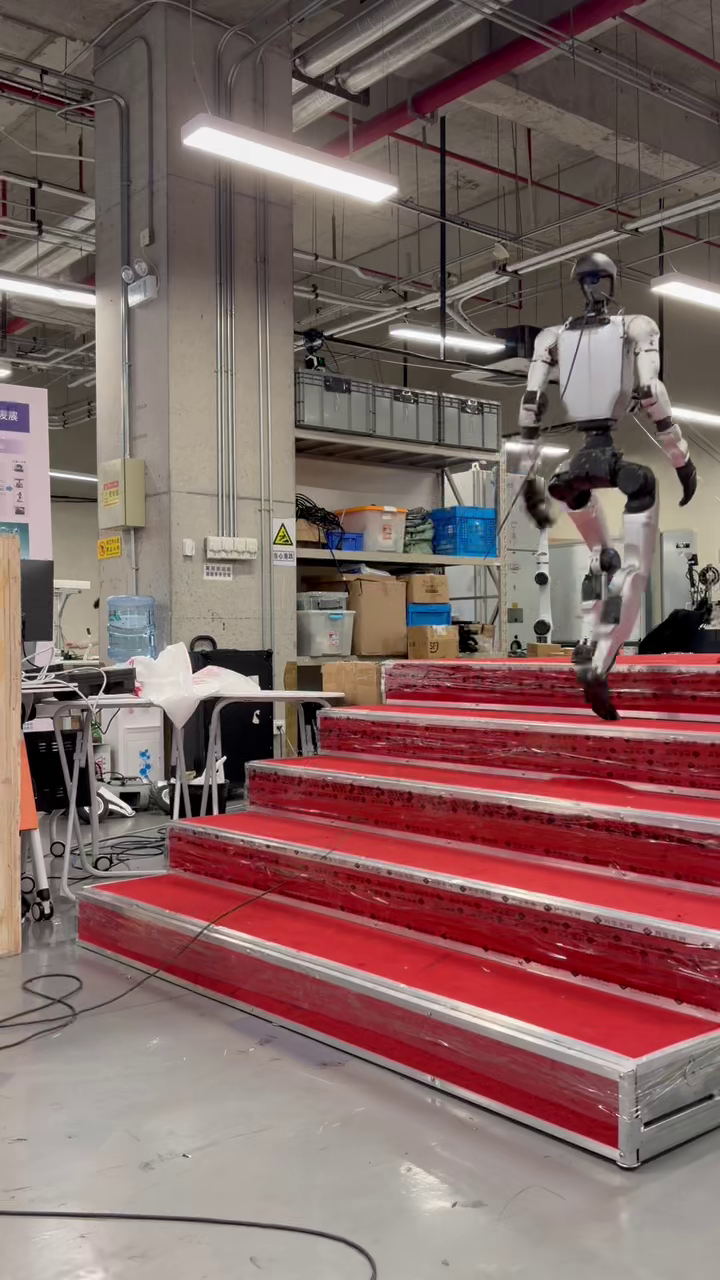}%
    \includegraphics[width=0.16\textwidth]{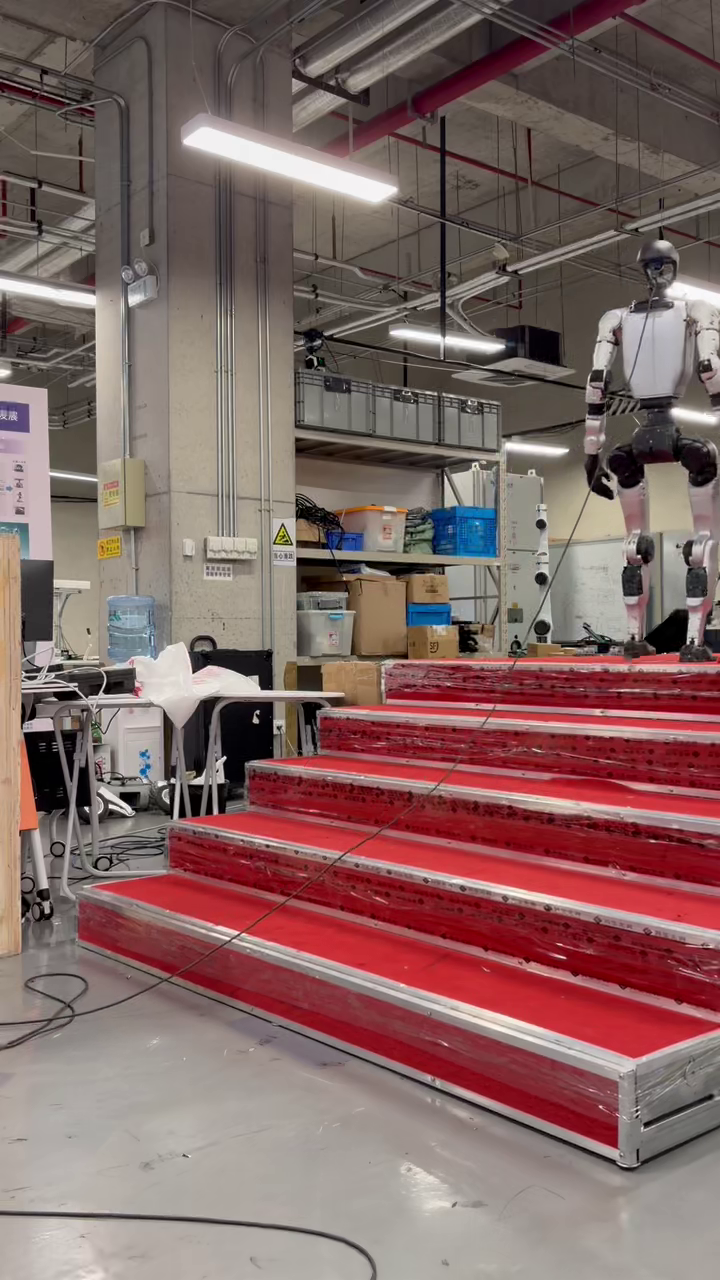}%
    \includegraphics[width=0.16\textwidth]{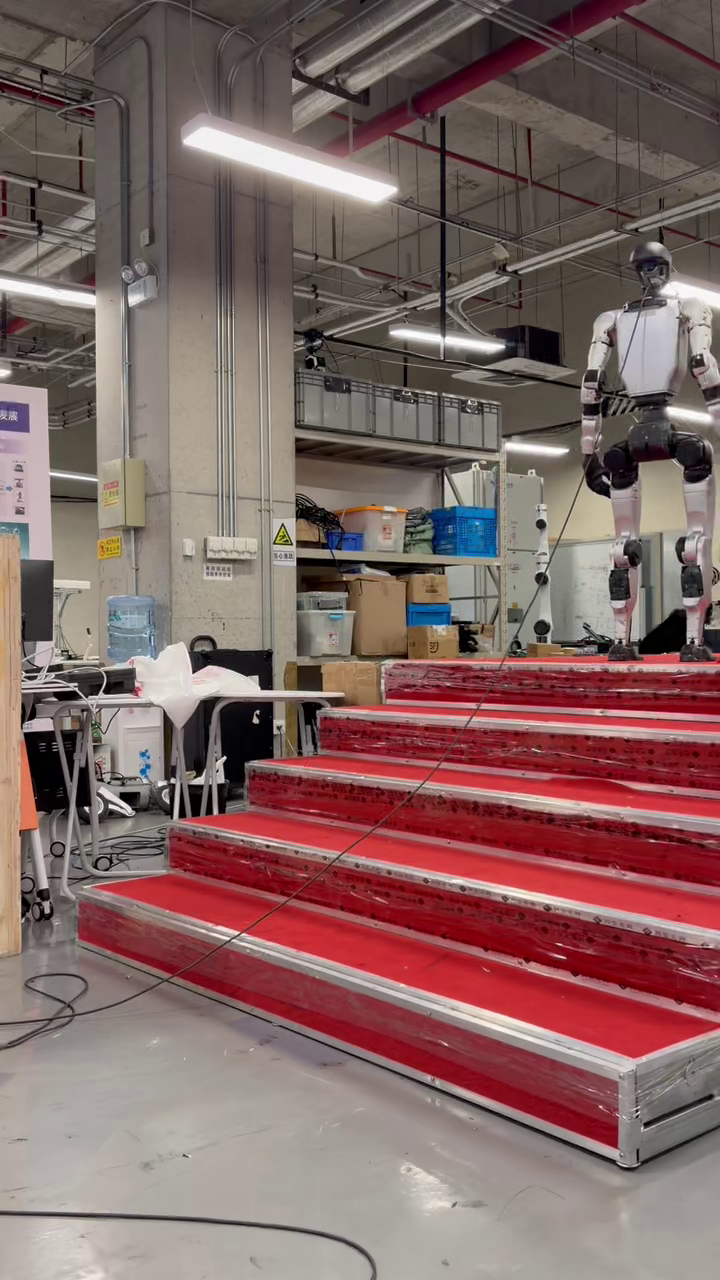}%
    \caption{Time-sequence key frames from real humanoid robot experiments in an indoor environment (left to right). The robot maintains stable locomotion during continuous indoor walking and motion transitions, demonstrating reliable real-world deployment of the proposed explicit stair geometry representation.}
    \label{fig:real_indoor_sequence}
\end{figure*}

\begin{figure*}[t]
    \centering
    \includegraphics[width=0.12\textwidth]{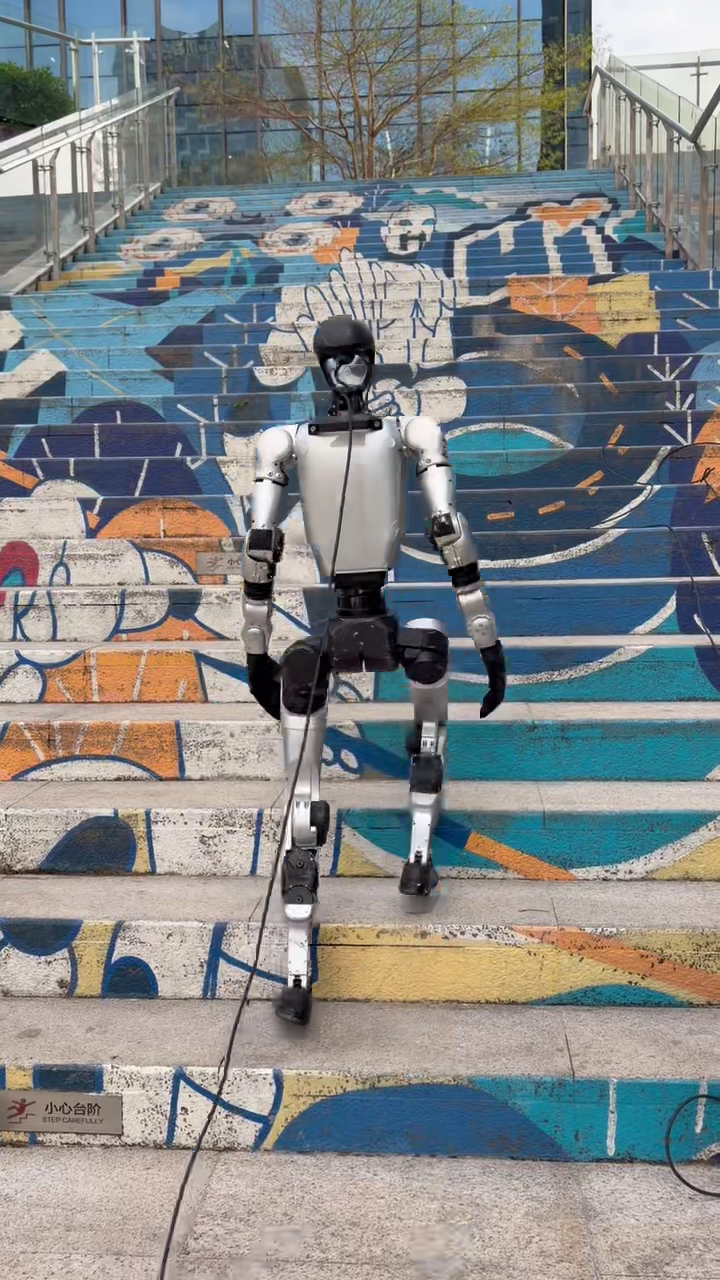}%
    \includegraphics[width=0.12\textwidth]{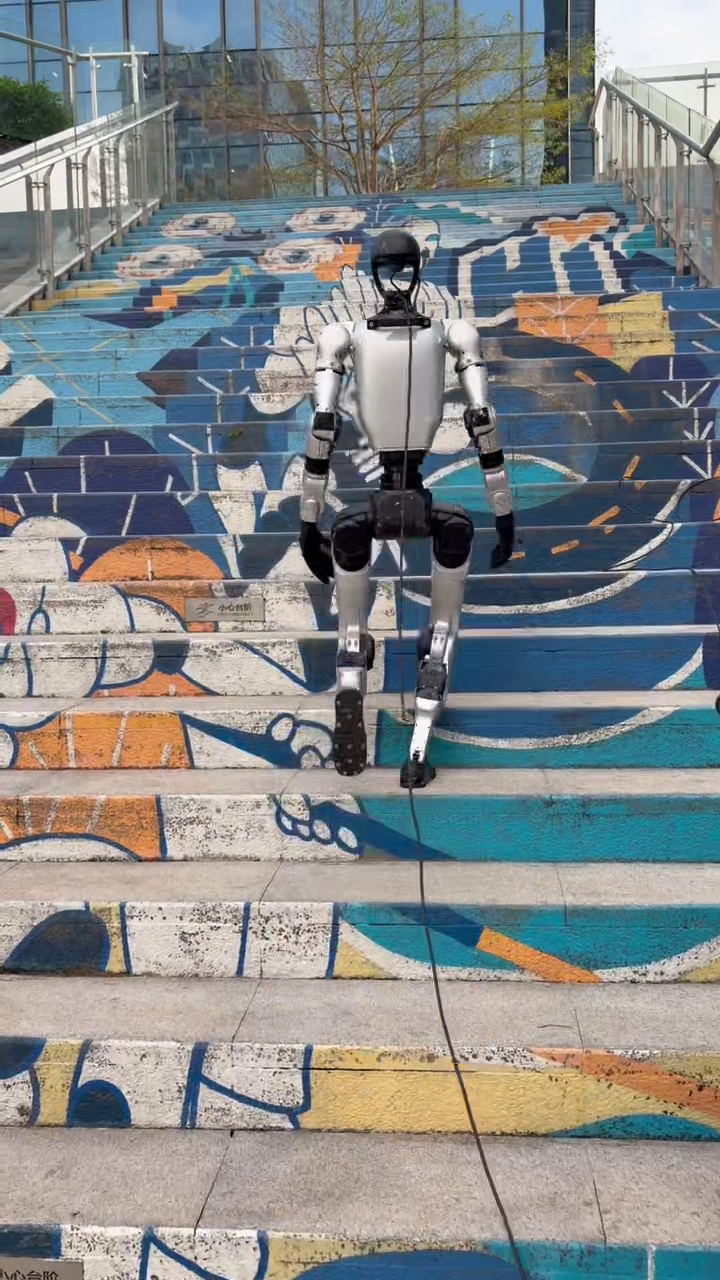}%
    \includegraphics[width=0.12\textwidth]{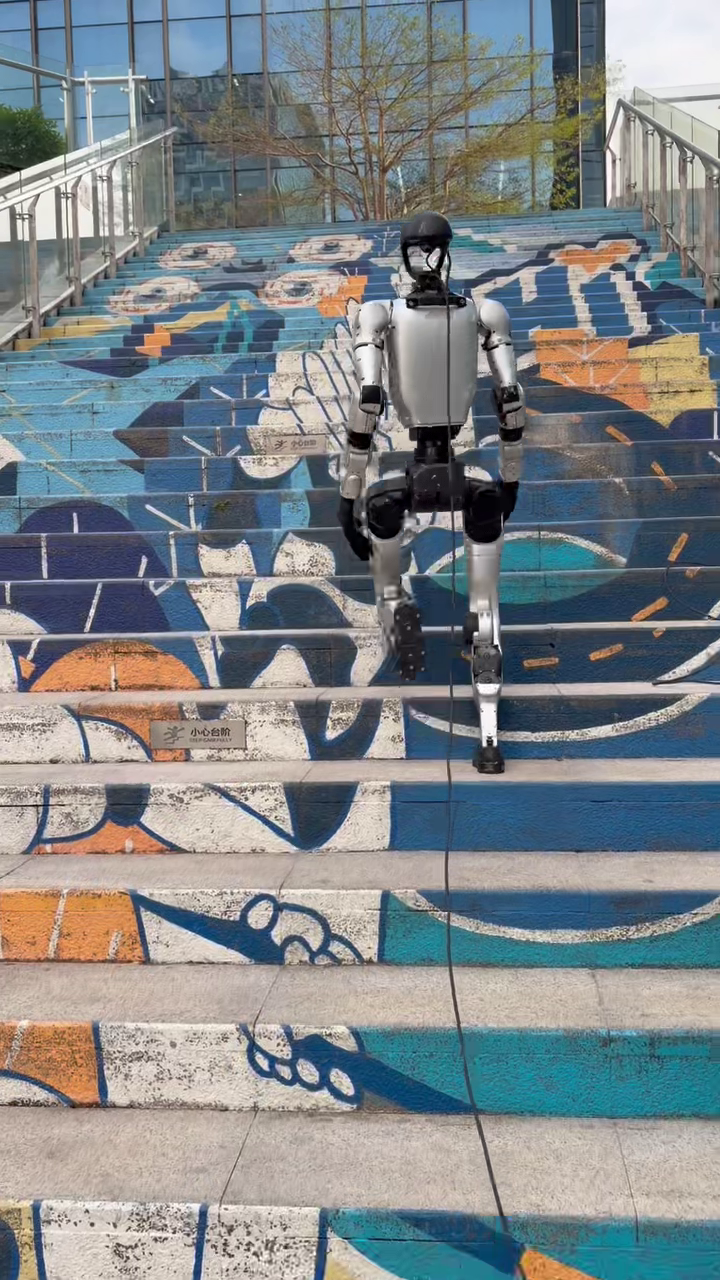}%
    \includegraphics[width=0.12\textwidth]{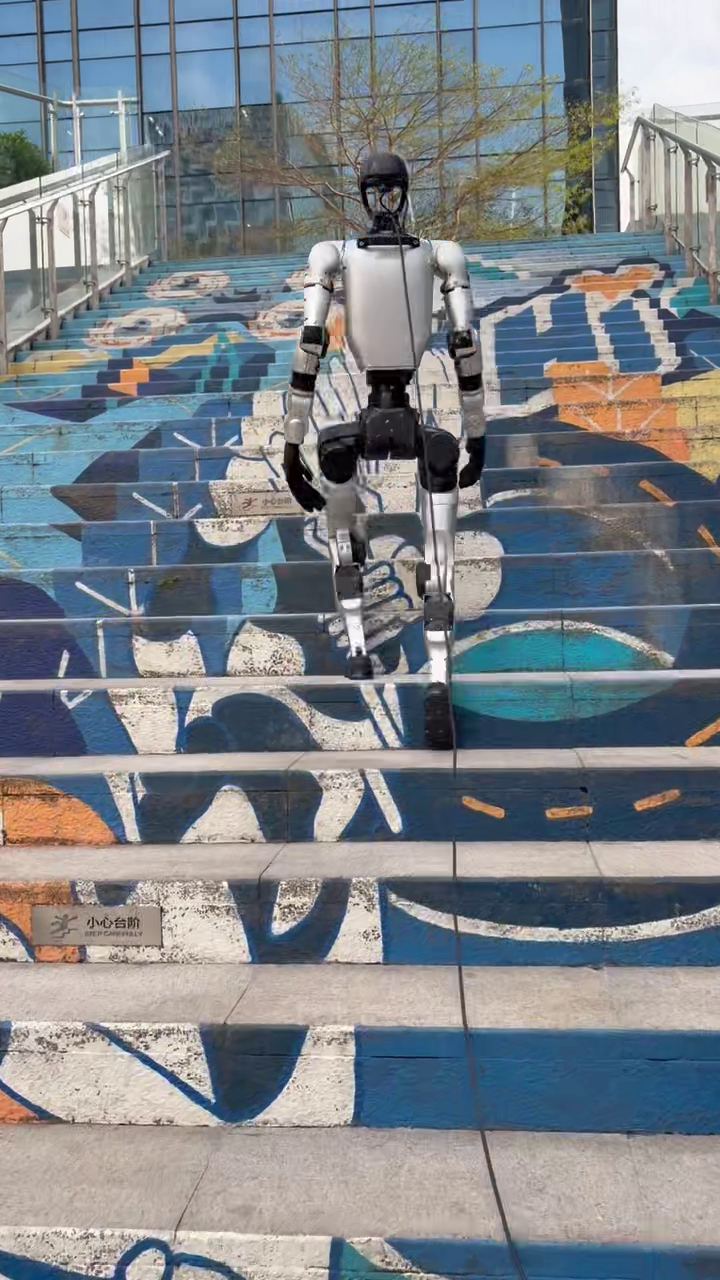}%
    \includegraphics[width=0.12\textwidth]{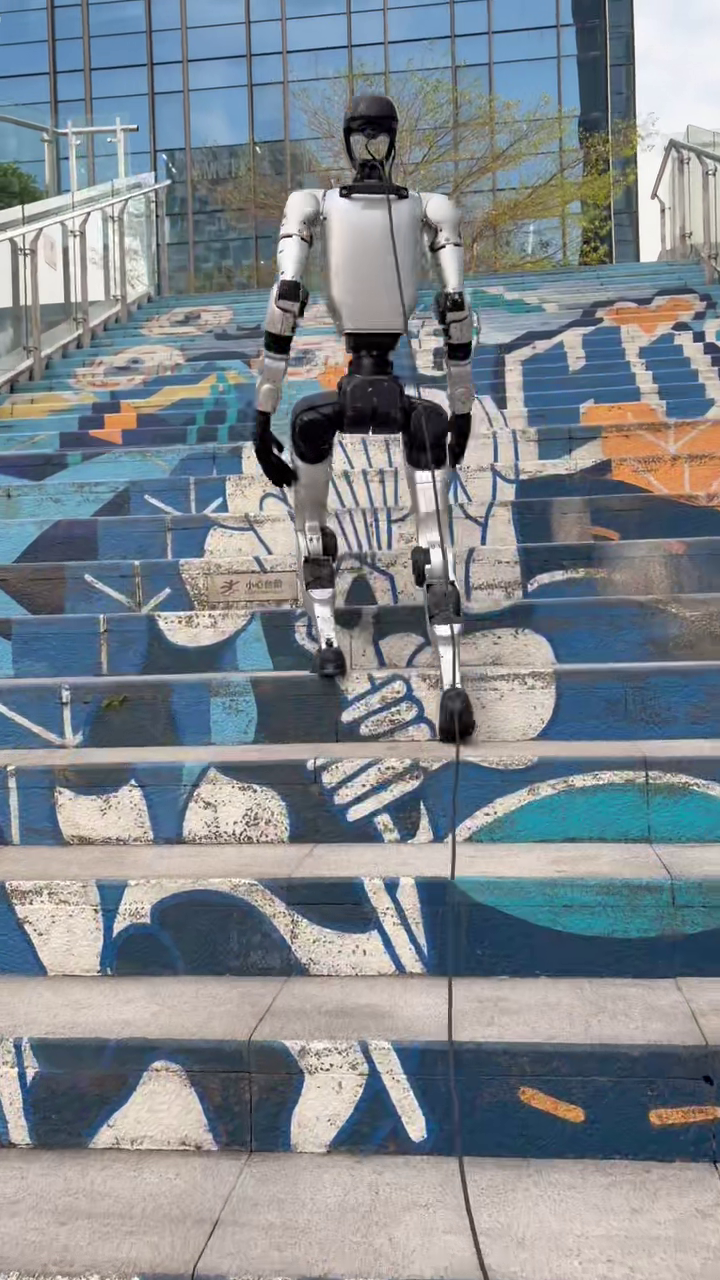}%
    \includegraphics[width=0.12\textwidth]{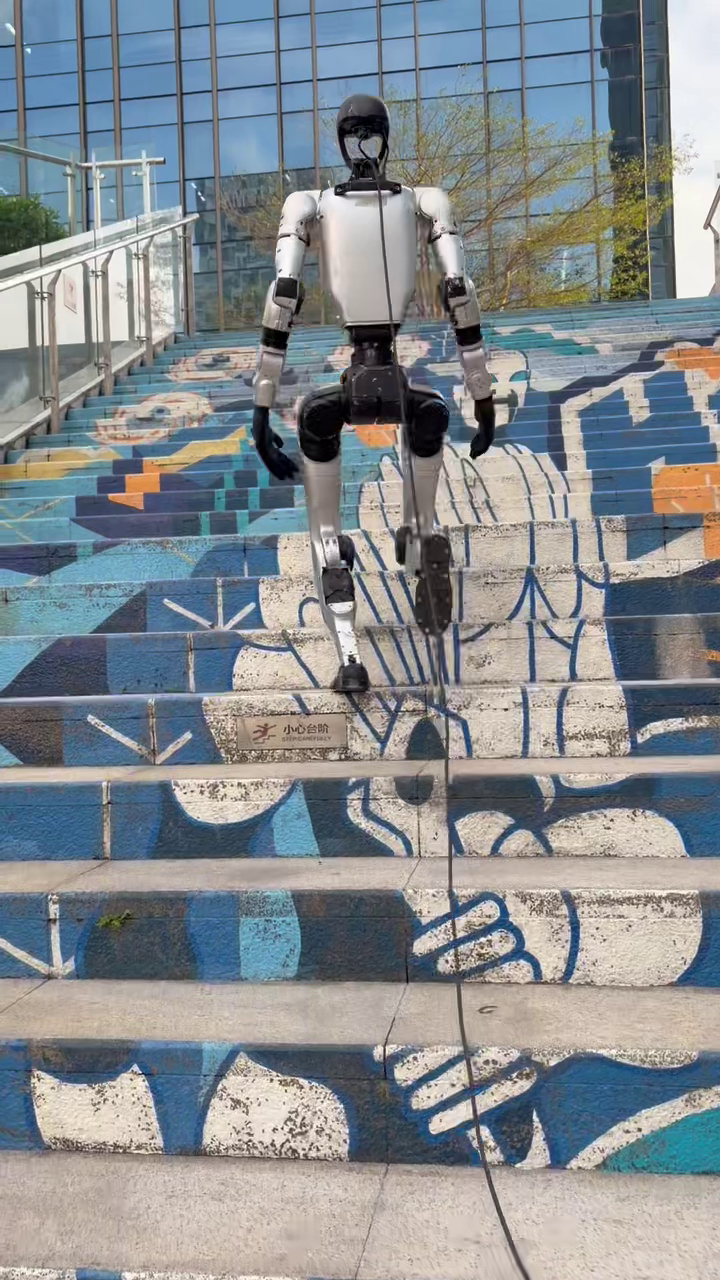}%
    \includegraphics[width=0.12\textwidth]{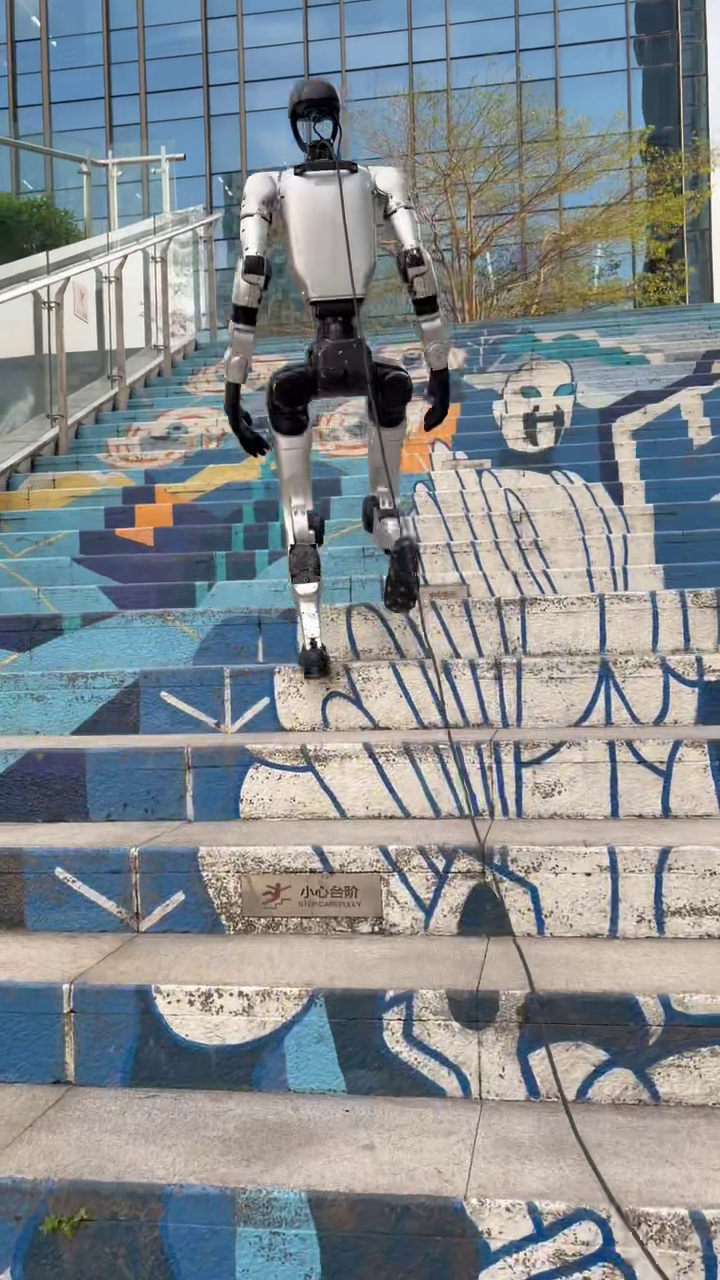}%
    \includegraphics[width=0.12\textwidth]{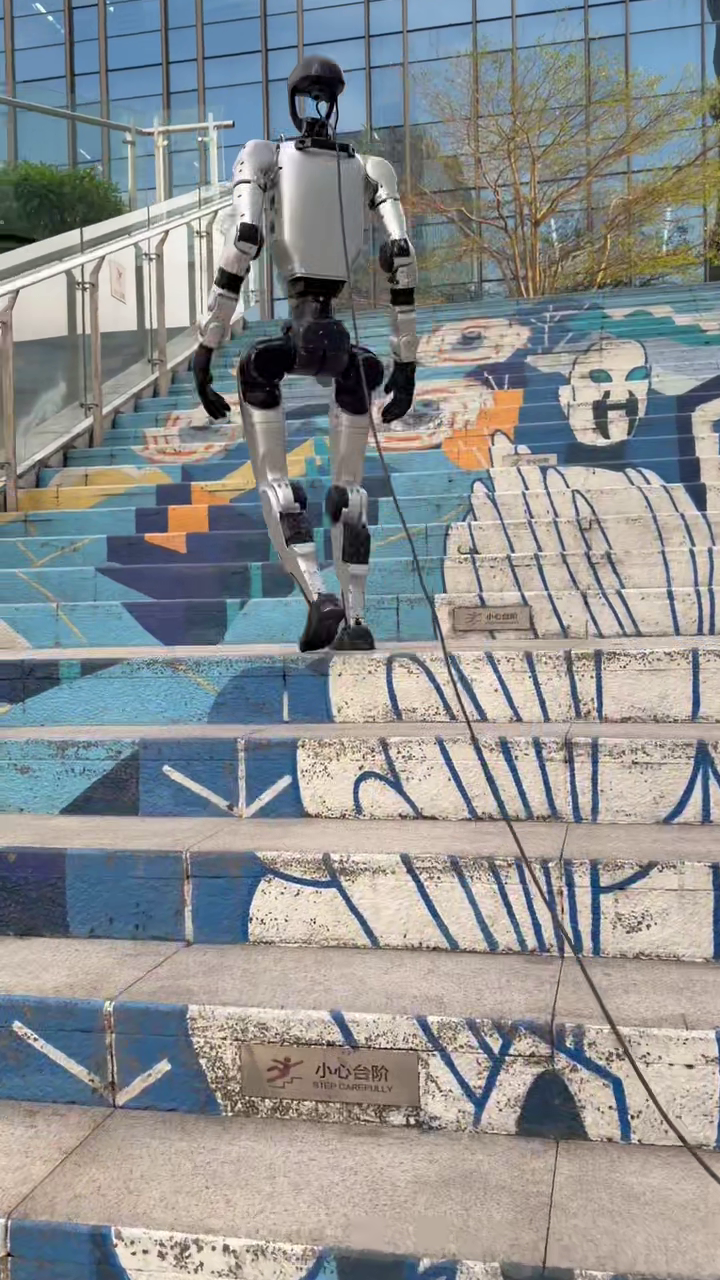}%
    \caption{Time-sequence key frames from real humanoid robot outdoor step-climbing experiments (1\,s--8\,s). The robot continuously ascends multiple steps with adaptive swing-foot clearance and coordinated body elevation, demonstrating robust real-world stair-climbing capability of the proposed explicit stair geometry representation.}
    \label{fig:real_outdoor_step_1s8s_mid}
\end{figure*}

Locomotion performance is evaluated using multiple complementary metrics:
linear and angular velocity tracking errors ($E_{\text{vel}}$, $E_{\text{ang}}$),
the maximum traversable terrain difficulty ($M_{\text{terrain}}$),
the normalized cumulative episode reward ($M_{\text{reward}}$),
and overall task success rate.
All results are reported as mean $\pm$ standard deviation over three random seeds.

As shown in Table~\ref{tab:baseline_comparison},
Blind-PPO exhibits large tracking errors and low success rate,
indicating that purely proprioceptive control is insufficient
for robust locomotion over structured stair environments.

HeightMap-PPO substantially improves terrain adaptability and success rate;
however, its performance remains sensitive to perception noise and discretization artifacts.

In contrast, the proposed explicit stair geometry representation achieves the lowest velocity tracking errors and the highest terrain difficulty,
reward, and success rate.
By explicitly encoding interpretable terrain parameters,
the policy can adapt its locomotion strategy according to underlying physical structure,
leading to more stable, robust, and sample-efficient learning across diverse stair geometries.

\subsubsection{Stair Geometry Estimation Accuracy}

To evaluate the robustness and reliability of the explicit stair geometry estimation module, we conduct quantitative analysis in both simulation and real-world environments.

\textbf{Simulation Evaluation.}
During training in simulation, Gaussian noise is injected into the depth input to mimic realistic sensor noise and perception uncertainty. 
For evaluation, we generate random stair configurations with varying step height, step depth, and yaw orientation. 
The trained perception module is evaluated over 1000 randomized episodes, and we compute the mean absolute error (MAE) for step height $h_{\text{step}}$, step depth $d_{\text{step}}$, and current yaw angle $\theta_{\text{yaw}}^{\text{current}}$, along with stair-state classification accuracy.

\begin{table}[t]
\centering
\scriptsize
\caption{Accuracy of explicit stair geometry estimation in simulation and real-world environments.
MAE denotes mean absolute error.}
\label{tab:geometry_accuracy}
\begin{tabular}{lcccc}
\toprule
Setting & MAE($h_{\text{step}}$) & MAE($d_{\text{step}}$) & MAE($\theta_{\text{yaw}}^{\text{current}}$) & State Acc (\%) \\
\midrule
Sim   & 0.6 cm & 0.7 cm & 1.8$^\circ$ & 99.1 \\
Real  & 0.9 cm & 1.1 cm & 3.9$^\circ$ & 97.3 \\
\bottomrule
\end{tabular}
\end{table}

\begin{table}[t]
\centering
\caption{Generalization performance across different stair heights. 
The first three heights are used during training, while higher stairs are unseen during testing.}
\label{tab:stair_generalization}
\begin{tabular}{lcc}
\toprule
\textbf{Stair Height} & \textbf{Vision-based (MoRE) \cite{wang2025moremixtureresidualexperts}} & \textbf{Ours} \\
\midrule
0.12m (Training) & 97.2\% & \textbf{99.7\%} \\
0.14m (Training) & 95.8\% & \textbf{99.4\%} \\
0.16m (Training) & 90.3\% & \textbf{97.5\%} \\
\midrule
0.18m (Testing)  & 88.5\% & \textbf{92.3\%} \\
0.20m (Testing)  & 78.8\% & \textbf{87.6\%} \\
0.22m (Testing)  & 69.4\% & \textbf{82.7\%} \\
\bottomrule
\end{tabular}
\end{table}

\begin{table}[t]
\centering
\caption{Success rate on irregular outdoor stairs.}
\label{tab:outdoor_stairs}
\begin{tabular}{lc}
\toprule
\textbf{Method} & \textbf{Success Rate} \\
\midrule
BlindPPO & 20\% \\
HeightMapPPO & 75\% \\
Vision-based (MoRE) & 72\% \\
Ours & \textbf{93\%} \\
\bottomrule
\end{tabular}
\end{table}






\textbf{Real-World Evaluation.}
In real-world experiments, we evaluate the perception module on two standard staircase configurations with known geometric parameters:
(i) step height 17 cm and depth 30 cm, and
(ii) step height 15 cm and depth 28 cm.
The humanoid robot collects approximately 200 depth observations under varying observation distances (0.5–2.0 m) and viewing angles.
Predicted geometric parameters are compared against manually measured ground-truth values, and estimation errors are computed accordingly.

The quantitative results are summarized in Table~\ref{tab:geometry_accuracy}. 
In simulation, the model achieves sub-centimeter accuracy for step height and depth estimation, and below $2^\circ$ error for yaw prediction. 
In real-world experiments, only moderate performance degradation is observed, maintaining centimeter-level geometric precision and over 97\% stair-state classification accuracy.

These results confirm that the explicit stair geometry representation provides stable and physically meaningful control-relevant parameters, enabling reliable geometry-conditioned locomotion across both simulated and real environments.

\subsubsection{Performance Trend Across Stair Heights}

We evaluate continuous stair-height generalization from 0.12 m to 0.22 m, where stair height corresponds to the geometric parameter $h_{\text{step}}$ estimated by our perception module. 
While both MoRE and our method perform well within the training distribution (0.12--0.16 m), performance degradation becomes evident in unseen heights.

As shown in Table~\ref{tab:stair_generalization}, MoRE drops from 90.3\% at 0.16 m to 69.4\% at 0.22 m, whereas our method maintains a higher success rate of 82.7\%.
The performance gap increases monotonically as stair height deviates from the training distribution, reaching a +13.3\% improvement at 0.22 m.

This trend indicates that the proposed explicit stair geometry representation enables physically grounded adaptation (e.g., increasing swing-foot clearance with stair height), rather than memorizing visual patterns.




\subsubsection{Generalization to Irregular Outdoor Stairs}
We further evaluate robustness on irregular outdoor staircases.

As summarized in Table~\ref{tab:outdoor_stairs}, our method achieves a 93\% success rate, significantly outperforming BlindPPO (20\%), HeightMapPPO (75\%), and MoRE (72\%).

Unlike heightmap-based methods that are sensitive to noise and vision-based policies affected by lighting variations, our explicit stair geometry conditioning provides robust structural cues, leading to improved deployment stability.



\subsection{Real-World Robot Experiments}

After training in simulation, the learned policy is directly deployed on a real humanoid robot platform without additional fine-tuning. Real-world experiments are conducted in both indoor and outdoor environments to evaluate robustness, generalization, and long-horizon performance.

\subsubsection{Indoor Stair-Climbing Experiments}

Indoor experiments are performed on a structured staircase consisting of five consecutive steps, representative of common indoor environments such as office buildings. The robot autonomously perceives the staircase, infers parameters and adjusts swing-foot clearance and body posture accordingly. As shown in Fig.~\ref{fig:real_indoor_sequence}, the robot successfully climbs all five steps in a continuous motion sequence without external support, demonstrating stable and smooth locomotion enabled by the proposed explicit stair geometry conditioning policy.

\subsubsection{Outdoor Stair-Climbing Experiments}

To further evaluate scalability and long-duration robustness, outdoor experiments are conducted on a large staircase consisting of 33 consecutive steps. This scenario introduces increased challenges due to prolonged execution time, accumulated state estimation errors, and environmental variability. Fig.~\ref{fig:real_outdoor_step_1s8s_mid} presents a time-sequence visualization extracted from 1\,s to 8\,s of the outdoor experiment, illustrating multiple step ascents with coordinated body elevation and adaptive swing-foot clearance. The robot successfully completes the entire staircase, highlighting the robustness and generalization capability of the proposed approach.

\section{Conclusion}

This paper presented an explicit stair geometry conditioning framework for robust humanoid stair climbing. 
By estimating compact and interpretable geometric parameters, including step height, step depth, and current yaw, and directly conditioning a PPO-based locomotion policy on these quantities, the proposed method enables proactive gait modulation without relying on implicit terrain embeddings or privileged information at deployment.

Simulation experiments demonstrated improved generalization across unseen stair heights compared to blind and heightmap-based baselines. 
The learned policy transferred directly to a real humanoid platform and achieved reliable indoor and outdoor stair traversal without additional fine-tuning. 
In challenging outdoor scenarios, the robot successfully ascended 33 consecutive steps, highlighting the robustness and practical deployability of explicit stair geometry conditioning.

\bibliographystyle{plain}
\bibliography{Embodided_AI}
	
\end{document}